\documentclass[sigconf]{acmart}
\AtBeginDocument{%
  \providecommand\BibTeX{{%
    \normalfont B\kern-0.5em{\scshape i\kern-0.25em b}\kern-0.8em\TeX}}}

\renewcommand\footnotetextcopyrightpermission[1]{}
\newcommand*{\affaddr}[1]{#1} 
\newcommand*{\affmark}[1][*]{\textsuperscript{#1}}
\renewcommand*{\email}[1]{\texttt{#1}}
\newcommand\blfootnote[1]{%
  \begingroup
  \renewcommand\thefootnote{}\footnote{#1}%
  \addtocounter{footnote}{-1}%
  \endgroup
}

\usepackage{threeparttable}
\usepackage{makecell}

\begin{document}

\title{Geometry Aligned Variational Transformer for Image-conditioned Layout Generation}

\author{%
Yunning Cao\affmark[1,2$\ast$], Ye Ma\affmark[2$\ast$], Min Zhou\affmark[2], Chuanbin Liu\affmark[1], Hongtao Xie\affmark[1$\dag$], Tiezheng Ge\affmark[2], Yuning Jiang\affmark[2]\\
\affaddr{\affmark[1]University of Science and Technology of China} \quad \affaddr{\affmark[2]Alibaba Group}\\
cynasd@mail.utsc.edu.cn, \{maye.my,yunqi.zm\}@alibaba-inc.com,\\ \{liucb92, htxie\}@ustc.edu.cn, \{tiezheng.gtz,mengzhu.jyn\}@alibaba-inc.com
}

\renewcommand{\shortauthors}{Yunning Cao, Ye Ma, Min Zhou, Chuanbin Liu, Hongtao Xie, Tiezheng Ge, Yuning Jiang}

\begin{abstract}
  Layout generation is a novel task in computer vision, which combines the challenges in both object localization and aesthetic appraisal, widely used in advertisements, posters, and slides design.
  An accurate and pleasant layout should consider both the intra-domain relationship within layout elements and the inter-domain relationship between layout elements and the image.
  However, most previous methods simply focus on image-content-agnostic layout generation, without leveraging the complex visual information from the image.
  To this end, we explore a novel paradigm entitled image-conditioned layout generation, which aims to add text overlays to an image in a semantically coherent manner.
  Specifically, we propose an Image-Conditioned Variational Transformer (ICVT) that autoregressively generates various layouts in an image.
  First, self-attention mechanism is adopted to model the contextual relationship within layout elements, while cross-attention mechanism is used to fuse the visual information of conditional images. Subsequently, we take them as building blocks of conditional variational autoencoder (CVAE), which demonstrates appealing diversity.
  Second, in order to alleviate the gap between layout elements domain and visual domain, we design a \textbf{Geometry Alignment} module, in which the geometric information of the image is aligned with the layout representation.
  In addition, we construct a large-scale advertisement poster layout designing dataset with delicate layout and saliency map annotations. Experimental results show that our model can adaptively generate layouts in the non-intrusive area of the image, resulting in a harmonious layout design. 
  
\blfootnote{$\ast$\;Equal contribution.\; \dag\,Corresponding author.}

\end{abstract}


\maketitle

\section{Introduction}
    \begin{figure}[htbp]
      \includegraphics[width=\linewidth]{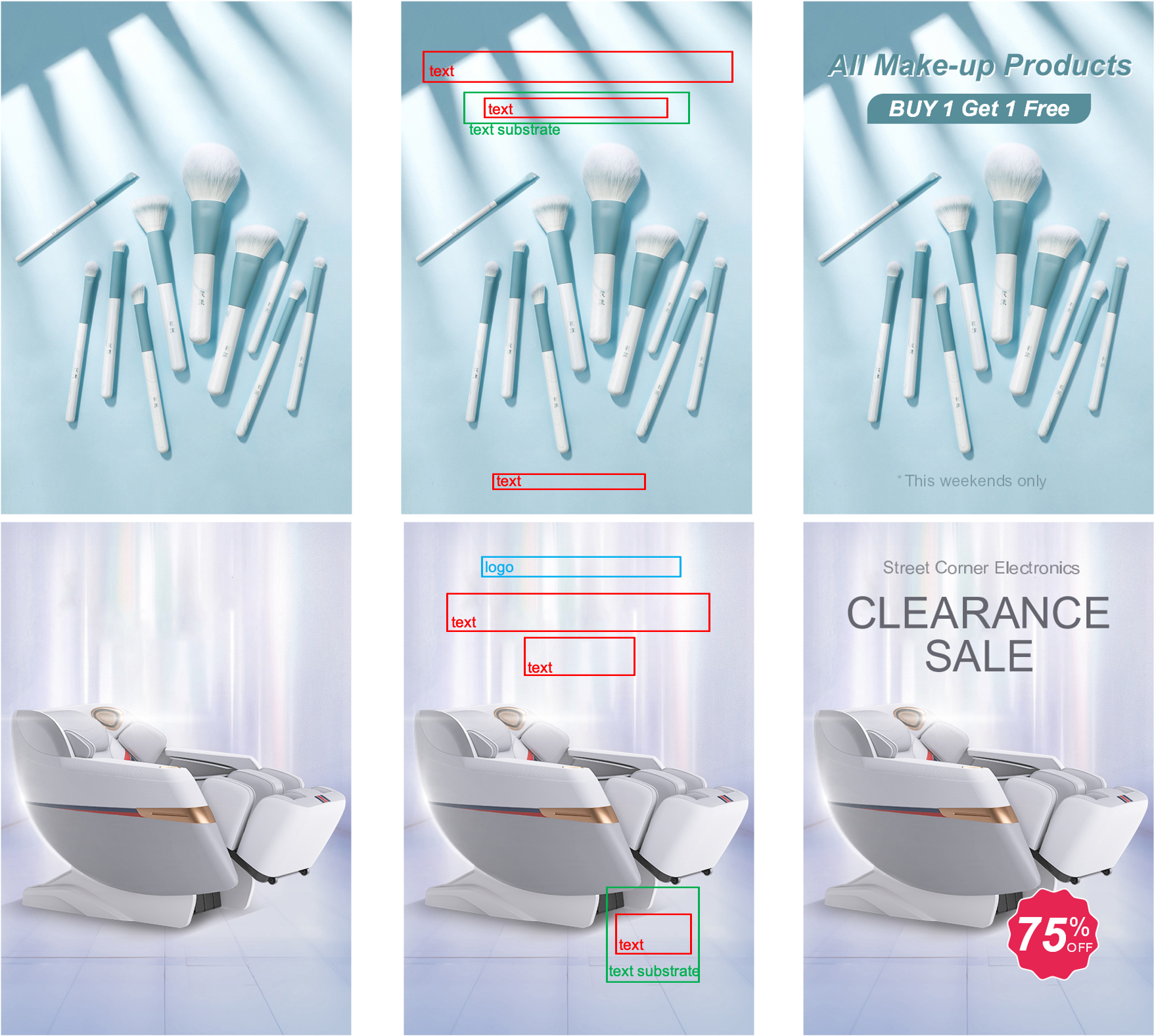}
      \caption{Examples of image-conditioned layout generation task. Given a background image with a subject presented (left), the layout consisting of multiple design elements needs to be generated (middle) following design principles. Hence, designers or even automatic rendering programs could utilize generated layouts to render the output images (right).}
      \label{fig:overview}
    \end{figure}
  Layout design aims at arranging design elements such as images, text and shapes in order to catch readers' attention and convey information in a visually appealing way. It is widely used in many  graphic design scenarios such as document typesetting, web page layout, poster, slides, billboards and user interface design, where a high aesthetic quality layout is crucial and fundamental. Therefore, the ability to automatically generate a high quality layout design without involving human designers is beneficial and has great value for future applications.
     
  Recently, layout design generation based on deep generative models such as Generative Adversarial Networks (GANs)~\cite{goodfellow2014generative} and Variational Autoencoders (VAEs)~\cite{vae} has attracted increasing interests. 
  Existing work has proved the effectiveness of deep learning to generate novel documents~\cite{contentGAN, layoutGAN,layoutTransformer}, user interfaces (UI)~\cite{attributeGAN,neuraldesign,vtn} and natural scenes~\cite{layoutTransformer,layoutVAE}. They utilize the ability of generative models to automatically generate possible and various layouts, either by directly predicting elements' bounding box locations~\cite{layoutVAE,neuraldesign,layoutTransformer,vtn}, or by synthesizing RGB images rendered from layouts~\cite{layoutGAN,attributeGAN,contentGAN}. 

  However, most prior papers only focus on the layout of foreground elements, ignoring the existence of background images. It may work well in some traditional situations such as document typesetting or web page layout, which usually presents a pure color background. But when it comes to the design of posters or advertisement billboards, whose background often contains the subject to be highlighted, the importance of background image arises if we want the generated design to be more natural and comparable with that of designers. Professional poster designers usually start from a given image with the subject centered and then consider where to put all other design elements, making the poster more native and natural. This differs from most existing work which only considers the relationship of design elements and fails to involve the semantic constraints between image context and generated layout elements.
  
  Thus, we claim that for layout design conditioned by an image background, the layout elements should be added to the image in a semantically coherent manner, such that the overall appearance of design elements and the image subject is harmonious and appealing. 
  We define this problem as \textit{image-conditioned layout generation}. 
  Note that some prior work mentioned conditional layout generation, which aims to generate layouts under conditions of either partial layout input~\cite{layoutTransformer} or the given number and types of elements\cite{layoutVAE}, which is different from our problem because they are image-content-agnostic methods.
  
  In this paper, we study the problem of image-conditioned layout generation. 
  As shown in Figure\ref{fig:overview}, there exists an appealing background image with other layout elements including text, text substrates, and logos. The positions of layout elements are determined by both image content and basic rules of layout design. For example, layouts should not occlude the salient subject in the image and text elements should not overlap with each other. Therefore, image-conditioned layout generation is more challenging than unconditional layout generation because it requires jointly modeling the intra-domain relationship within layout elements and inter-domain relationship between layout elements and the background image. 
  
  To solve this problem, we first parameterize layout elements to their categorical and geometric parameters (bounding boxes with classes). Then, we project each element of the layout into an embedding vector and represent the whole layout as a sequence of vectors. Thus, we formulate the problem as a conditional sequence autoregressive prediction problem. Consequently, we propose a transformer based conditional variational autoencoder (CVAE) to jointly model the layout sequence and conditional image. We adopt self-attention layers to depict the intra-domain relationship within layout elements and cross-attention layers to learn the inter-domain relationship between image and layout elements. Then we take them as building blocks of CVAE. VAE framework plays an important role in modeling real data distribution with its latent space, allowing us to generate various layouts by latent space sampling. 
  
  Furthermore, as the most unique thing of image-conditioned layouts compared to unconditional ones is the relationship between the visual domain (the background image) and geometric parameter domain (bounding box locations), feature alignment turns out to be a fundamental and critical problem to solve. Specifically, we need to extract the geometric information of the image and project it into the same domain with parameterized layout elements. There exists prior art trying to solve it by involving geometry enhancement module to transformers~\cite{meng2021conditional,liu2021dab}. However, they only focus on geometry prior of input queries, without considering the relationship between images and queries. Therefore, we propose a novel geometry alignment module to extract the image geometry information and align it with the layout elements. We divide the image into grids and represent each grid as its bounding boxes, which is aligned to the representation of layout bounding box. After that, it is much easier to model the geometry relationship between image and layout bounding boxes.
  
  To validate our proposed method, we construct a large-scale advertisement poster design layout dataset, consisting of 117,624 poster images designed by professional designers, annotated with rich in-image layout annotations and saliency map of the background product subject. Extensive experiments on the proposed dataset demonstrate the effectiveness of the proposed ICVT model. Compared with previous unconditioned layout generation models, our ICVT model shows much better results in image-conditioned layout generation, achieving the state-of-the-art performance.
  
  Our main contributions can be summarized as follows:
  \begin{itemize}
      \item We propose the Image-Conditional Variational Transformer (ICVT) for visual-textual layout design generation, which is one of the first deep learning based models for image-conditioned layout generation to the best of our knowledge. 
      \item We propose a novel geometry alignment module to enhance inter-domain feature fusion by decoupling and projecting image geometry information into the same domain of layout elements.
      \item We construct a large-scale advertisement poster design layout dataset, consisting of 117,624 poster images designed by designers with rich annotations of layouts.
  \end{itemize}

\section{Related Work}

  Recently, layout generation is increasingly attracting interest in the research community. 
  We summarize existing work according to how they represent layout elements and divide them into two types: 
  One is rendering layout elements into visual images, which we refer to as the visual domain.
  The other is using geometry and attribute parameters to express any layout elements, which we refer to as the geometric parameter domain.
  
  \textbf{Visual domain.}
  To the best of our knowledge, LayoutGAN~\cite{layoutGAN} is the first to apply generative adversarial network (GAN) to layout generation. 
  To leverage the CNN discriminator's advantage of distinguishing visual patterns, they propose a novel differentiable wireframe rendering layer to rasterize structured data of layout elements into wireframe images. 
  Therefore, the layout can be directly optimized in the visual domain.
  However, they simply parameterize all elements and render them into wireframe images, without any visual information.
  Following LayoutGAN, Attribute-Aware LayoutGAN~\cite{attributeGAN} adds editable attribute conditions such as element area, aspect ratio, and order. 
  Although involving image elements, they simply represent the image as a bounding box and attribute descriptions, without involving any visual information from the image. 
  Besides,~\cite{contentGAN} directly takes an image-based representation for layouts, without involving any layout parameters (i.e. bounding box). 
  Then they train a VAE-GAN conditioned on images, keywords, and attributes of the layout, which generates content-aware graphic design layouts. 
   
  Visual representation has the potential for cross-domain modeling (e.g. jointly modeling image and parameterized layouts), as it directly converts the layout parameters to the visual domain, naturally achieving domain alignment. However, representing layouts in the visual domain could not scale well when generating layouts of bigger size and resolution, where the computational cost and sophisticated post-processing algorithms such as contour or smoothing need to be considered. 

  \textbf{Geometric parameter domain.}
  Parameterized layout representation (e.g. bounding box) is a much more popular way. 
  LayoutVAE~\cite{layoutVAE} first proposes an autoregressive model based on VAE~\cite{vae} to generate various layouts. It generates layouts from a set of labels, i.e. the categories of layout elements. 
  However, it is built with LSTMs, which suffers from the problem of long sequence modeling, resulting in the difficulty of modeling a large number of layout elements. 
  As transformer~\cite{attn_is_all} shows a great power of sequence modeling, the following work gradually adopt the transformer architectures instead of LSTMs. 
  LayoutTransformer~\cite{layoutTransformer} addresses the layout generation problem using an autoregressive language model based on transformers. 
  They find that self-attention can learn relationships between layout elements, yielding competitive layout quality.
  Furthermore, to increase the diversity of generated layouts, Variational Transformer Network (VTN)~\cite{vtn} is proposed. 
  VTN integrates an autoregressive transformer decoder into the VAE framework to generate various layouts by sampling from latent space.
  Neural Design Network (NDN)~\cite{neuraldesign} involves Graph neural network~\cite{scarselli2008graph} to model the relations among layout elements.
  Compared to the visual domain, parameterized layout representation is much more flexible. However, the domain gap between parameterized layouts and images obstructs cross-domain modeling.
  
  \textbf{Geometry enhancement} is an important method for enhancing the performance of cross-attention. Detection Transformer (DETR)~\cite{carion2020end} directly fuses queries with visual features without geometry enhancement, which results in an extremely slow training convergence. To solve the problem, conditional DETR~\cite{meng2021conditional} designs 2D positional embedding $(x, y)$ as geometry prior for the queries, which increases the speed of convergence and accuracy. Furthermore, DAB-DETR~\cite{liu2021dab} proposes a 4D positional embedding $(x,y,w,h)$ for queries to learn dynamic geometric prior. However, the above methods focus on enhancing geometry information for queries, without considering the alignment of geometry information between queries and image. 
  Besides, in the image caption, geometry enhancement is used to help model image spatial information~\cite{guo2020normalized, zhang2021rstnet}. Geometry-aware model~\cite{guo2020normalized} calculates the relative geometry relation among different objects to benefit visual reasoning. As the development of transformer, RSTN~\cite{zhang2021rstnet} utilizes a similar method to model relative geometry relation among grid features.
  However, these methods focus on enhancing geometric information in the visual domain, without cross-domain alignment.

\section{Approach}
  \begin{figure*}[htbp]
    \includegraphics[width=0.8\textwidth]{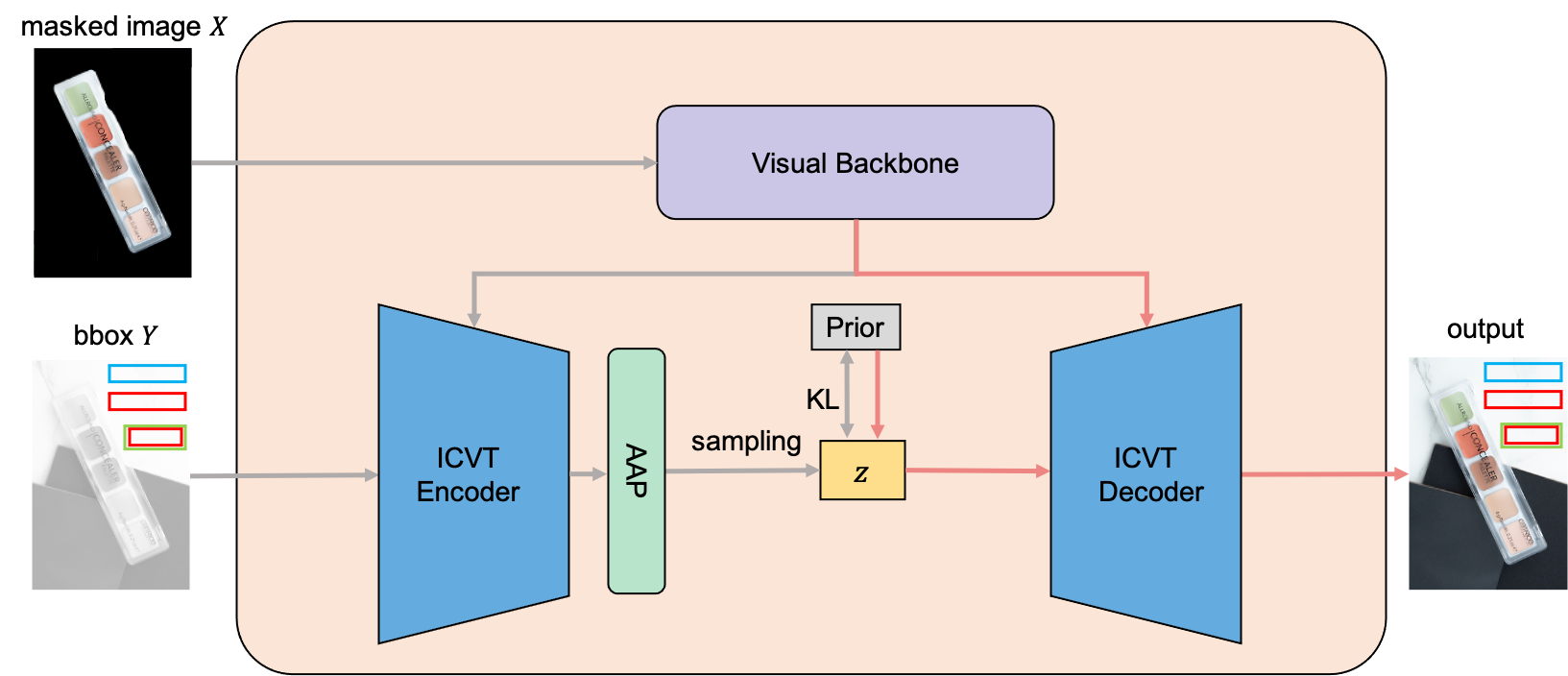}
    \caption{Overall architecture of Image-Conditioned Variational Transformer (ICVT). During training (both gray and red arrows), masked image $\mathbf{X}$ is put into the visual backbone and fed into ICVT encoder and decoder as a condition, while bounding boxes $\mathbf{Y}$ is put into ICVT encoder-decoder. Attention Average Pooling (AAP) aggregates the encoder output. The model objective is the combination of reconstruction loss and KL loss. During inference (red arrows only), we sample latent representations $\mathbf{z}$ from prior and autoregressively generate layouts.}
    \label{fig:ICVT}
  \end{figure*}
  \subsection{Problem Formulation}
    We are interested in jointly modeling the intra-domain relationships within layout elements and the inter-domain relationships between layout elements and the image, and furthermore generating diverse and appealing layouts on the given image. The problem can be defined as follows:
    
    Given a dataset of image-layout pair $\mathcal{D}=\left\{\mathbf{X}_{i}, \mathbf{Y}_{i}\right\}_{i=1}^{|\mathcal{D}|}$, the $\mathbf{X}_i$ represents image and $\mathbf{Y}_i=[\mathbf{Y}_{i}^1, \mathbf{Y}_{i}^2,\dots, \mathbf{Y}_{i}^T]$ represents $T$ layout elements on the corresponding image. 
    Specifically, a layout element can be defined with its category and a bounding box $ (c_i^j, x_i^j, y_i^j, w_i^j, h_i^j)$. We project the above attributes to the same $d$-dimentional space and further concatenate them into a layout vector $\mathbf{Y}_i^j$, where $\mathbf{Y}_i^j\in \mathbb{R}^{5d}$. Besides, we use a latent variable $\mathbf{z}$ to control the prediction and involve diversity.
    
    Subsequently, our problem can be formulated as modeling the distribution of layout elements $\mathbf{Y}$ conditioned on the image $\mathbf{X}$ and a random vector $\mathbf{z}$ (latent vector in CVAE). $\hat{\mathbf{Y}}$ is the generated layouts:
    \begin{equation}
        \hat{\mathbf{Y}}=\arg\max_{\mathbf{Y}} P(\mathbf{Y}\mid \mathbf{z}, \mathbf{X})
    \end{equation}
    
  \subsection{Conditional Variational Autoencoders}
    The problem of image-conditioned layout generation requires generating various layouts from a single input image, which can be formulated as one-to-many mapping. 
    Consequently, we adopt Conditional Variational Autoencoder~\cite{sohn2015learning, walker2016uncertain} as it naturally models the one-to-many mapping from input to output, with no need for an explicitly specified structure of the output distribution. 
    
    The goal of CVAE is to approximate the conditional data distribution $p_{\theta} (\mathbf{Y} \mid \mathbf{X})$ by maximizing the conditional data log-likelihood $\mathbb{E}_{\mathbf{X}, \mathbf{Y} \sim \mathcal{D}}\left[\log p_{\theta} (\mathbf{Y} \mid \mathbf{X})\right]$. Considering the intractable problem of posterior inference, a $\phi$-parameterized encoder is involved to approximate $p_{\theta} (\mathbf{z} \mid \mathbf{X}, \mathbf{Y}) \propto p_{\theta} (\mathbf{Y} \mid \mathbf{z}, \mathbf{X}) p (\mathbf{z}\mid \mathbf{X})$ with a variational distribution $q_{\phi} (\mathbf{z}\mid \mathbf{X}, \mathbf{Y})$. Variational inference is employed for CVAE learning with the following evidence lower bound (ELBO): 
    \begin{equation}
      \mathcal{L} (\theta, \phi)=\mathbb{E}_{q_{\phi} (\mathbf{z} \mid \mathbf{X}, \mathbf{Y})}\left[\log \left (p_{\theta} (\mathbf{Y} \mid \mathbf{z}, \mathbf{X})\right)\right]-K L\left (q_{\phi} (\mathbf{z} \mid \mathbf{X}, \mathbf{Y}) \| p(\mathbf{z} \mid \mathbf{X})\right)
      \label{eq:ELBO}
    \end{equation}
    where $p(\mathbf{z}\mid \mathbf{X})$ is prior distribution and $q_{\phi} (\mathbf{z} \mid \mathbf{X}, \mathbf{Y})$ is posteriror distribution.

  \subsection{ICVT Architecture}
    As shown in Figure, our proposed ICVT model contains three main components, including a visual backbone to extract image features and a transformer-based encoder for generating latent code and a transformer decoder as a conditional generator.
    \subsubsection{\textbf{Visual Backbone}}
      The visual backbone learns visual features from the image, and use them as a condition to guide the process of layout generation. 
      Specifically, we utilize a vision transformer (ViT)~\cite{vit} as it has a better relation modeling ability. 
      Given an image $\mathbf{X}\in \mathbb{R}^{H\times W\times C}$, we reshape it into a sequence of flattened 2D $P\times P$ patches. 
      After that we feed them into a ViT model, generating a visual feature $f\in \mathbb{R}^{l\times d}$, where $L=\frac{HW}{P^2}$ is the length of the patch sequence, $d$ is the embedding dimension of the visual backbone. 
      Besides, to align the embedding dimension of the visual backbone and that of variational transformer, we add a transformer encoder after the visual backbone.
    \subsubsection{\textbf{ICVT Encoder}}
      The ICVT encoder models the posterior distribution $q_\phi (\mathbf{z}\mid \mathbf{X}, \mathbf{Y})$ in the general CVAE framework. 
      Thus, we need to encode both the image $\mathbf{X}$ and layouts $\mathbf{Y}$ into a joint conditional posterior distribution, which is a challenging problem that distinguishes our model from other unconditional layout generation models like VTN~\cite{vtn} and LayoutVAE~\cite{layoutVAE}. 
      There are several common fusion methods like directly concatenating the visual features with the input layout embeddings, aggregating visual features into a single vector and fusing it with the latent vector $\mathbf{z}$, etc.
      However, after some comparisons, cross-attention mechanism comes to be the best choice to fuse the visual features with layout features, because it can not only keep the most spatial information of the image but also naturally model the inter-domain relationship between visual features and layout features. 
      
      Based on the above reasons, we implement our ICVT encoder with a standard transformer decoder~\cite{attn_is_all}, which consists of self-attention layer, cross-attention layer, and feed-forward network (FFN). 
      In this module, the layout features are taken as the input sequence. As the coordinate of layout bounding boxes already contains spatial position information, we do not add positional encoding to layout features.
      While the visual features are taken as keys and values of cross-attention layers, with positional encoding added to keys. After that, we get the joint representation of layouts and images, which is a sequence of vectors with its length equal to the input layout sequence. 
      After that, to thoroughly utilize the information in the joint representation, we design an attention average pooling (AAP) layer to aggregate the sequence of vectors into a single vector.
      Specifically, the AAP layer is implemented with a self-attention layer, where the query $Q\in\mathbb{R}^{1\times D}$ is a single learnable vector while the keys and values $K=V\in \mathbb{R}^{L\times D}$ are the sequence of vectors from the ICVT encoder. Next, we use the aggregated vector to model the posterior distribution.
      
      The posterior distribution $q_\phi (\mathbf{z}\mid \mathbf{X}, \mathbf{Y})$ is parameterized as a multi-variant normal distribution with diagonal covariance matrix, i.e. $N (\mu, \sigma^2\mathbf{I})$, where $\mathbf{I}$ is identity matrix. We calculated the mean vector $\mu$ and $\log\sigma$ vector by projecting the aggregated vector with linear layers. As for prior distribution, we compare standard normal distribution $N (0,\mathbf{I})$ with learnable prior and empirically take non-learned prior the reason will be discussed in the ablation study.
      Besides, to implement backpropagation training for encoder, reparameterization trick~\cite{cvae} is used to allow gradient passing through Gaussian sampling.
      
      During the training process, the latent code $\mathbf{z}$ is sampled from posterior distribution while in inference process, the latent code is sampled directly from prior distribution.
      
    \subsubsection{\textbf{Autoregressive Decoder}}
      The ICVT decoder $p_{\theta} (\mathbf{Y} \mid \mathbf{z}, \mathbf{X})$ shares the same architecture with the encoder. 
      The only difference is that the decoder is performed as an autoregressive generator, i.e. $p_{\theta} (\mathbf{Y} \mid \mathbf{z}, \mathbf{X})=\prod_{i=1}^l p_{\theta} (\mathbf{Y_{i}} \mid \mathbf{Y_{i-1}}, \mathbf{z}, \mathbf{X})$, where $l$ is the number of bounding boxes in a layout. We take the latent code $\mathbf{z}$ as the begin of sequence (BOS) token. The generated tokens are projected by five independent fully-connected layers to predict $cls, x, y, w, h$ separately.
      
  \subsection{\textbf{Geometry Alignment}}\label{subsec:GA}
    In our model, cross-attention modules play an important role in jointly modeling the relationship between the layouts and the conditional image. 
    However, the layout features are embeddings of class and geometric parameters (bounding boxes) while the conditions are image visual features, there is a domain gap between them. 
    We find that vanilla cross-attention is not good enough to solve the problem of domain gap.
    For example, the dot product of the query $Q$ and the key $K$ is meaningless if $Q$ and $K$ come from different representation spaces.
    
\begin{figure}[htbp]
      \centering
      \includegraphics[width=0.9\linewidth]{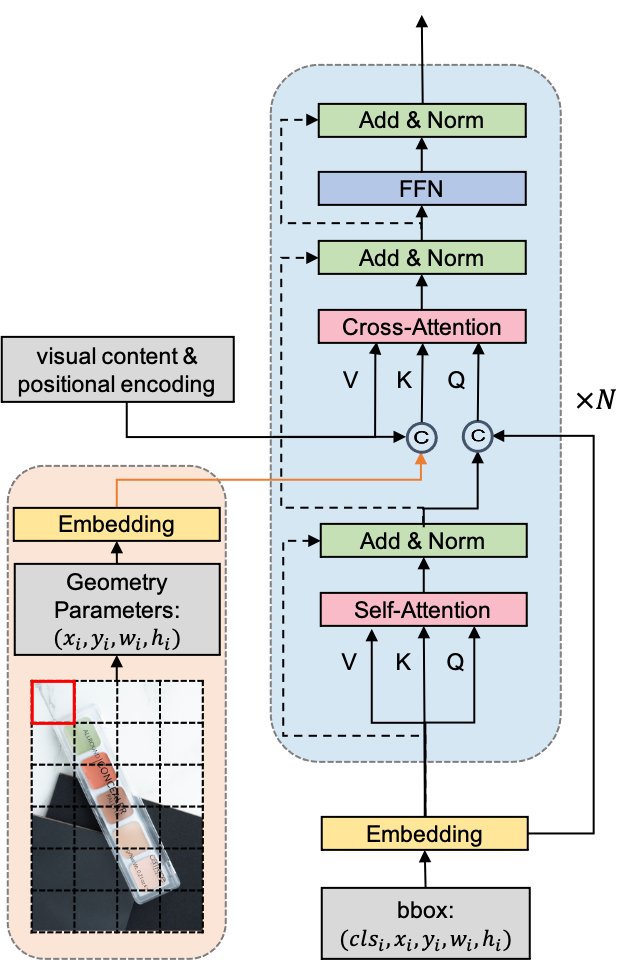}
      \caption{Geometry Alignment Module (orange) and Transformer Block (blue). We take the concat fusion method as an example.}
      \label{fig:GA}
    \end{figure}

    To solve the problem of domain gap, we propose a Geometry Alignment (GA) Module as shown in Figure\ref{fig:GA}. Considering the visual features (taken as $K$ and $V$) contain no explicit geometric information, our GA module calculates geometric parameters $ (x_j, y_j, w_j, h_j)$ for each patch feature of the image and projects them into the embedding space $\mathbb{R}^D$. In the following, we propose three geometry fusion methods between visual features and layout features.
    
    \subsubsection{\textbf{Adding}}
    A natural fusion method is simply adding geometry information $Q_g(K_g)$ into content information $Q_c(K_c)$ as shown in the following Equation \eqref{eq:add}, where $Q$ represents layout features, $K$ and $V$ are both visual features.
    \begin{equation}
      Q = Q_c + Q_g,  \quad K = K_c + K_g, \quad V = V_c
      \label{eq:add}
    \end{equation}
    After that, we can calculate the cross-attention through the following Equation \eqref{eq:add_cross}, where $\sigma$ represents the softmax operator. In the second and third line of Equation \eqref{eq:add_cross}, we expand the expression and divide it into three parts including the content term $Q_cK_c^T$, the geometry term $\color{red}{G=Q_gK_g^T}$ and cross term $\color{blue}{CrossTerm=Q_c K_g^T+Q_g K_c^T}$.
    \begin{align}
        \begin{aligned}
          \text{CrossAttn} &= \sigma\left (\frac{Q K^T}{\sqrt{d_k}} \right)V_c\\
					&= \sigma\left (\frac{Q_c K_c^T+{\color{red}Q_g K_g^T}+{\color{blue}Q_c K_g^T+Q_g K_c^T}}{\sqrt{d_k}}\right)V_c\\
					&= \sigma\left (\frac{Q_c K_c^T+{\color{red}G}+\color{blue}CrossTerm}{\sqrt{d_k}}\right)V_c
		\end{aligned}
		\label{eq:add_cross}
    \end{align}
    \subsubsection{\textbf{Concat}}
    Another fusion way is concatenation. Based on the above equation, we observe that the content and geometry information are coupled in the cross terms, which is redundant. 
    Thus, we propose to decouple the content and geometry information by concatenating them along the embedding dimension as shown in Equation \eqref{eq:concat}:
    \begin{equation}
      Q = \text{concat} (Q_c , Q_g),  \quad K = \text{concat} (K_c , K_g), \quad V = V_c
      \label{eq:concat}
    \end{equation}
    Putting these concatenated features into cross attention achieves to decouple the content and geometry information, so that content query and geometry query focus on content and geometry attention weights respectively.
    \begin{align}
        \begin{aligned}
          \text{CrossAttn} &= \sigma\left (\frac{Q K^T}{\sqrt{d_k}} \right)V_c\\
		     		&= \sigma\left (\frac{Q_c K_c^T+\color{red}Q_g K_g^T}{\sqrt{d_k}}\right)V_c\\
					&= \sigma\left (\frac{Q_c K_c^T+\color{red}G}{\sqrt{d_k}}\right)V_c
		\end{aligned}
    \end{align}
    \subsubsection{\textbf{Manually designed geometry relation}}
    Furthermore, inspired by the above analysis, we propose to replace the geometry term $\color{red}G$ with any manually designed geometry term, which explicitly calculates the relative geometry relations. For example, we imitate the computation of region geometry features in~\cite{guo2020normalized, zhang2021rstnet} to obtain the relative geometry features between layouts and image patches. Our methods differ from them because they calculate relations within image features while we calculate the inter-relationship between layout features and image features.
    \begin{align}
          r_{i j}=\left (\begin{array}{c}
          \log \left (\frac{\left|c x_{i}-c x_{j}\right|}{w_{i}}\right) \\
          \log \left (\frac{\left|c y_{i}-c y_{j}\right|}{h_{i}}\right) \\
          \log \left (\frac{w_{i}}{w_{j}}\right) \\
          \log \left (\frac{h_{i}}{h_{j}}\right)
          \end{array}\right) \\
          {\color{red}G_{i j}}=\operatorname{ReLU}\left (w_{g}^{T}FC\left (r_{i j}\right)\right)
    \end{align}
    where $i$ and $j$ refer to layout bounding boxes and image grids respectively, $FC$ is a fully-connected layer, and $w_g$ is a learnable weight matrix. 
  \subsection{ICVT Optimization}
    Since our ICVT is a CVAE model, we take the standard ELBO defined in Equation \eqref{eq:ELBO} as the optimization objective.
    
    However, with an autoregressive decoder, the optimization suffers from the famous posterior collapse problem~\cite{bowman2016generating}.
    Concretely, during the training, the decoder tends to ignore the information in the latent code and depends on the masked input sequence. At the same time, $\mathbf{z}$ learned from encoder perfectly matches the prior, transmitting no information to decoder.
    To solve the problem, we follow the $\beta$-VAE~\cite{higgins2016beta} objective in Equation \eqref{eq:betaVAE}, which adds a weight $\beta$ to KL term in Equation\eqref{eq:ELBO},  :
    \begin{equation}
      \mathcal{L} (\theta, \phi)=\mathbb{E}_{q_{\phi} (\mathbf{z} \mid \mathbf{X}, \mathbf{Y})}\left[\log \left (p_{\theta} (\mathbf{Y} \mid \mathbf{z}, \mathbf{X})\right)\right]-\beta K L\left (q_{\phi} (\mathbf{z} \mid \mathbf{X}, \mathbf{Y}) \| p(\mathbf{z} \mid \mathbf{X})\right)
      \label{eq:betaVAE}
    \end{equation}
    and adopt a cyclic annealing schedule~\cite{fu-etal-2019-cyclical} for $\beta$. Specifically, we set $\beta$ close to zero at the early steps of a cycle, and linearly increase it till it reaches the predefined $\beta$ weight. The above schedules are repeated periodically.
    
    An intuitive explanation for the method is as follows. In the early training period, with $\beta$ closing to zero, the model degenerates to a VAE model and learns an informative latent representation $\mathbf{z}$. When the $\beta$ increases to non-negligible, the earlier learned $\mathbf{z}$ conveys enough information to decoder, which alleviates the posterior collapse.

\section{Experiments}
  
  \subsection{Dataset}
  Training our ICVT model requires a large and diverse dataset with background image and design layout pairs. However, existing public layout datasets~\cite{zhong2019publaynet, coco} only focus on layout element locations ignoring background image and hence not applicable to our problem. Consequently, we build a large-scale advertisement poster layout dataset. It consists of 117,624 visual-textual advertisement poster images designed by professional designers, with both image background of the product subject and rich in-image design layout annotations. Bounding boxes of design texts, text substrates, and logos are annotated by outsourcing tasks. Besides, image saliency map is processed by a state-of-the-art matting model and attached to each image. We split our dataset into a training set and a validation set in 9:1 ratio and all hyper-parameters are selected on the validation set.  
  
  For testing, we construct a new test set with 166 background images, similar to training images while without any layout elements (i.e. images before design). Similarly, image saliency maps are processed with the same matting model and fed into models for inference.
  
  In the following experiments, we use boxes of different colors to identify layout elements of different classes, with red for text, green for text substrates, and blue for logos.
  
  \subsection{Implementation Details}
    Our ICVT is a transformer model with a ViT~\cite{vit} visual backbone and an encoder-decoder structure based on standard transformer components described in~\cite{attn_is_all}. 
    For visual backbone, we adopt ViT-S/32 i.e. ViT small model with a patch size of $32\times 32$ model initialized with ImageNet-pretrained parameters. The ICVT decoder is stacked with 4 transformer decoder layers, with the head number set to 8. The embedding dimension of each component of bounding boxes (class, x, y, w, h) is 96, summing up to 480 as the model dimension. The dimension of FFN is set to 2048 and the dropout probability is 0.1. The ICVT encoder shares the same architecture as decoder.
    
    We use AdamW~\cite{adamw} to optimize the model, where we set encoder-decoder's learning rate to $5\times 10^{-5}$, while a smaller learning rate $1\times 10^{-5}$ for the pretrained visual backbone. 
    Besides, weight decay is set to $1\times 10^{-2}$ and batch size is set to 128. 
    We train the model for two cycles with a cyclic annealing schedule for $\beta$ parameter. Concretely, the cyclic schedule is formulated as follows:
    \begin{equation}
      \begin{aligned}
        \beta_{t}&=\left\{
        \begin{array}{cl}
          0.001, & 0\le t \le T/2 \\
          f (t), & T/2 < t <3T/4\\
          0.3, & 3T/4 \le t < T
        \end{array} \quad\right.
      \end{aligned}
      \label{eq:beta}
    \end{equation}
    where $t$ is iteration numbers, $T$ is the total iteration numbers of a period, $f (t)$ is a linear function determined by two points $ (T/2, 0.001)$ and $ (3T/4, 0.3)$. 
    
    Also, we use data augmentation including color jitter for the input image and random flip for both image and layouts. 
    Finally, our ICVT is implemented in PyTorch framework~\cite{pytorch} and is trained with 16 NVIDIA V100 GPUs.
    \begin{table*}[!htbp]
      \caption{Ablation study on Geometry Alignment}
      \label{tab:ablation}
      \begin{tabular}{llcccccc}
        \toprule
        Method & Geometry Embedding & Output Rate $\uparrow$ & Overlap $\downarrow$ & Alignment $\downarrow$ & Occlusion $\downarrow$\\
        \midrule
        \texttt{baseline} & --- & 97.8\% & 0.033 & 0.015 & 0.184 \\
        \texttt{baseline(w/o PE)} & --- & 97.6\% & \textbf{0.030} & 0.018 & 0.264 \\
        \texttt{adding} & \texttt{learned} & \textbf{98.8\%} & 0.036 & 0.014 & 0.181 \\
        \texttt{adding} & \texttt{sine} & 98.2\% & 0.031	& 0.014	& 0.175 \\
        \texttt{concat} & \texttt{learned} & 98.5\% & 0.055 & 0.014 & 0.161 \\
        \texttt{concat} & \texttt{sine} & \textbf{98.8\%} & 0.046 & \textbf{0.013} & \textbf{0.154} \\
        \texttt{manually}& --- & 98.4\% & 0.048 & 0.016 & 0.165\\
        \bottomrule
      \end{tabular}
  \end{table*}
  
  \subsection{Evaluation Metrics}
    Layout design follows some basic rules. 
    For example, layout elements should be aligned with each other as possible and should not overlap with each other.
    And when it comes to image-conditioned layout generation, layout elements should not occlude with the salient object in background image.
    Evaluating the quality of the image-conditioned layouts involves two main aspects---intra-bounding-boxes relationship and relationship between image and bounding boxes. 
    First, we follow previous work~\cite{layoutGAN, neuraldesign} to take alignment and overlap as metrics for evaluating the layout bounding box quality. 
    Then, different from unconditional layout generation, in order to measure the relationship between the background image and bounding boxes, we design a new metric, occlusion, which is defined as the overlap area of background saliency map and design element bounding boxes over the total bounding box area. A lower occlusion metric leads to a better layout.
    Besides, we also evaluate the global visual quality by calculating Fréchet Inception Distance (FID)~\cite{fid_heusel2017gans} following the setting mentioned by NDN~\cite{neuraldesign}.
    Last but not least, we calculate model output rate (inference samples with at least one valid box over all samples) as indicators for the consistency and stability of models.
    
  \subsection{Quantitative results}
  \subsubsection{\textbf{Ablation study}}
  As shown in Table \ref{tab:ablation}, we perform ablation studies to investigate the effectiveness of different model structures. We present an ICVT model without geometry alignment as a baseline. 
  First, we study the importance of image positional encoding by completely removing it from the baseline model. We find that the occlusion rate significantly increases from 0.184 to 0.264, while other metrics almost remain unchanged. It demonstrates that the positional encoding is critical for learning spatial information of the image. 
  
  Then, we study the effect of the three fusion methods mentioned in Section \ref{subsec:GA}, along with different embedding methods for geometry information. Experiments show that concatenation is a better fusion method than addition. And for both methods, sine geometry embedding performs slightly better than learned geometry embedding in occlusion rate.
  Manually designed geometry relation also performs well, with occlusion rate slightly higher than the concat method. However, we do not adopt this method because we find that the calculation of relative geometry relation and the following fully-connected layer is computationally expensive.

  Finally, we study the performance of learned prior and non-learned prior on the best ICVT model chosen from the above ablation study.  
  As shown in Table \ref{tab:ablation_prior}, we find that the non-learned prior performs better in output rate and occlusion rate, with nearly the same overlap and alignment performance. This result is unusual as in most conditional variational autoencoders, learned prior including conditional information leads to better results. One possible reason is that our task depends highly on spatial information of the conditional image, 
 while the learned prior could harm the cross-attention of queries and keys. Consequently, we use standard normal prior in the following experiments. 
  \begin{table}[htbp]
  \caption{Ablation study on prior distribution. }
  \label{tab:ablation_prior}
  \resizebox{\linewidth}{!}{
  \begin{tabular}{lcccccc}
    \toprule
    Prior & Output Rate $\uparrow$ & Overlap $\downarrow$ & Alignment $\downarrow$ & Occlusion $\downarrow$\\
    \midrule
    \texttt{non-learned} & \textbf{98.8\%} & 0.046 & \textbf{0.013} & \textbf{0.154}\\
    \texttt{learned}&  97.6\% & \textbf{0.045} & 0.014 & 0.165\\
    \bottomrule
  \end{tabular}}
  \end{table}

  \subsubsection{\textbf{Comparison with prior art}}
  Based on the above ablation studies, we choose the ICVT model with concatenated sine geometry embedding to perform the following experiments. 
  
  To compare with previous models, we reproduce the content-aware GAN~\cite{contentGAN}, Layout Transformer~\cite{layoutTransformer}, and VTN~\cite{vtn} to train them on our proposed dataset. 
  Note that there is no image input for VTN and Layout Transformer while an extracted image embedding is fed to content-aware GAN.
  As shown in Table \ref{tab:comparison}, compared with VTN and Layout Transformer, our model shows similar performance on output rate, overlap, and alignment metrics. 
  However, VTN and Layout Transformer yield a much higher occlusion rate than our ICVT model due to the lack of visual information, which also leads to the higher FID. As for content-aware GAN, although it involves visual information, our ICVT model performs better on alignment, occlusion rate, and FID. It is worth mentioning that the overlap metric of content-aware GAN is the smallest because it generates rendered images instead of parameters of geometry boxes. Those coordinates are obtained via contour algorithm and border smoothing, which makes it a leading position on the metric of alignment.
  
  In all, the result demonstrates that our ICVT model is state-of-the-art in jointly modeling layout elements and visual information.
  
  \begin{table}[htbp]
  \caption{Quantitative comparison with the prior art. }
  \label{tab:comparison}
  \resizebox{\linewidth}{!}{
  \begin{tabular}{lcccccc}
    \toprule
    Method & Output Rate $\uparrow$ & Overlap $\downarrow$ & Alignment $\downarrow$ & Occlusion $\downarrow$ & FID $\downarrow$\\
    \midrule
    \texttt{contentGAN~\cite{contentGAN}} &  \textbf{100.0\%} & \textbf{0.016} & 0.015 & 0.334 & 85.81\\
    \texttt{LayoutTransformer~\cite{layoutTransformer}} & \textbf{100.0\%} & 0.038 & 0.017 & 0.252 & 71.58\\
    \texttt{VTN~\cite{vtn}} & 98.4\% & 0.045 & 0.015 & 0.254 & 79.77\\
    \texttt{ICVT(ours)}& 98.8\% & 0.046 & \textbf{0.013} & \textbf{0.154} & \textbf{62.04}\\
    \bottomrule
  \end{tabular}}
  \end{table}
  
  \subsection{Qualitative results}
  \subsubsection{\textbf{Image-conditioned layout generation and completion.}} 
\begin{figure}[htbp]
      \centering
      \includegraphics[width=\linewidth]{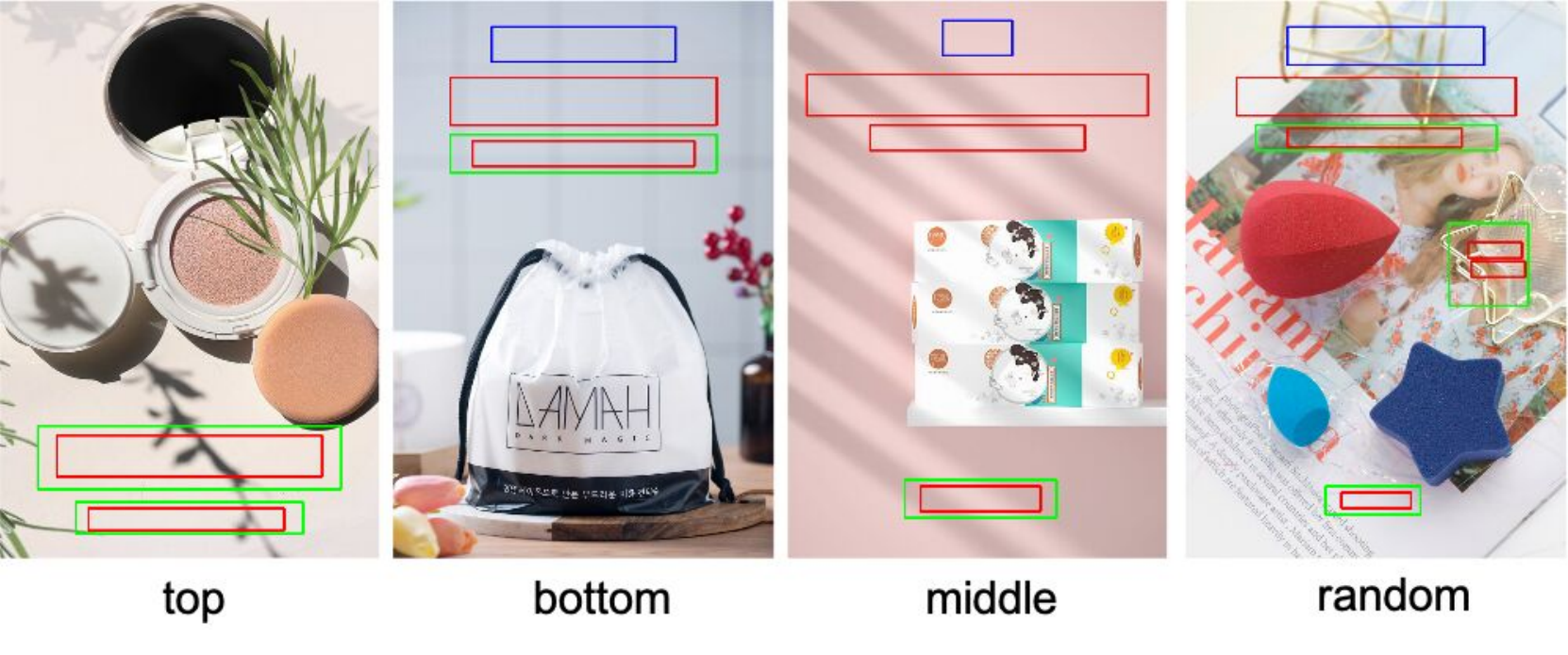}
      \caption{Image-conditioned layout generation. From left to right, the salient object is in the top, bottom, middle and random distributed on the image. Our model can adaptively generate layouts on different images properly.}
      \label{fig:generation}
  \end{figure}

\begin{figure}[htbp]
      \centering
      \includegraphics[width=\linewidth]{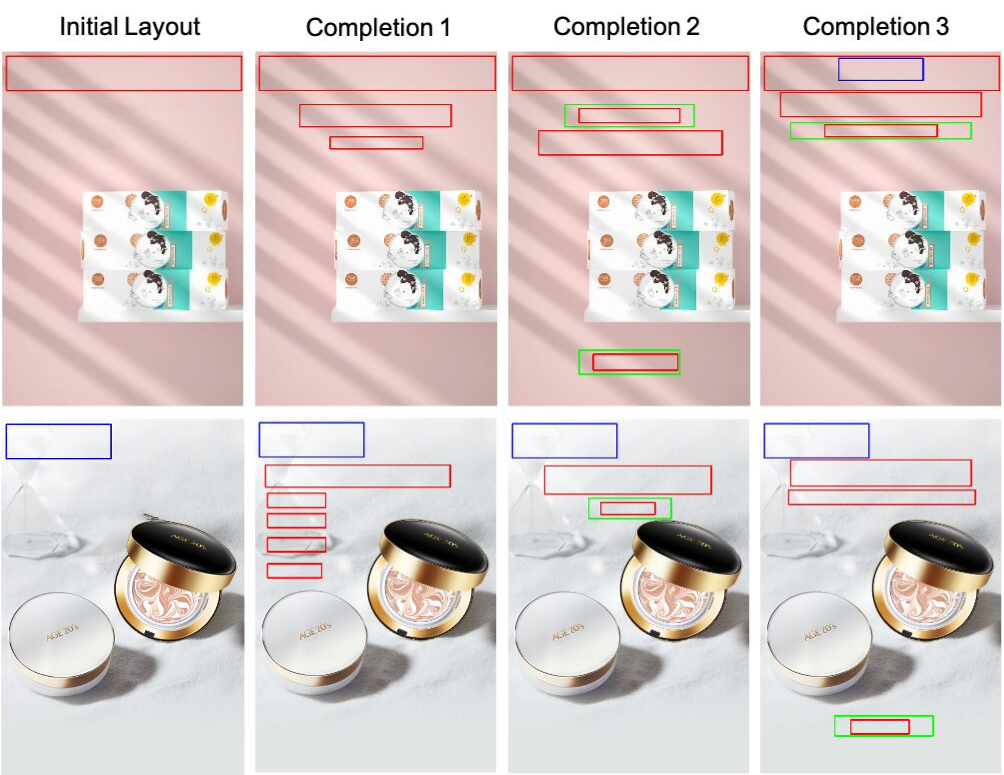}
      \caption{Multiple Completions from the same initial element layout.}
      \label{fig:completion}
  \end{figure}

We perform both image-condition layout generation and completion as qualitative results. 
  As shown in Figure \ref{fig:generation}, We choose images (masked with saliency map) with different salient object locations to show that our ICVT model can adaptively generate layouts on the non-invasive regions of different images, resulting in a proper and beautiful layout design. 
  Furthermore, autoregressive model can naturally complete partial layouts. As shown in Figure \ref{fig:completion}, taking a partial layout as input, our ICVT model can complete layouts based on given initial elements.
    
  \subsubsection{\textbf{Comparison with prior art}}
\begin{figure}[!htbp]
      \centering
      \includegraphics[width=\linewidth]{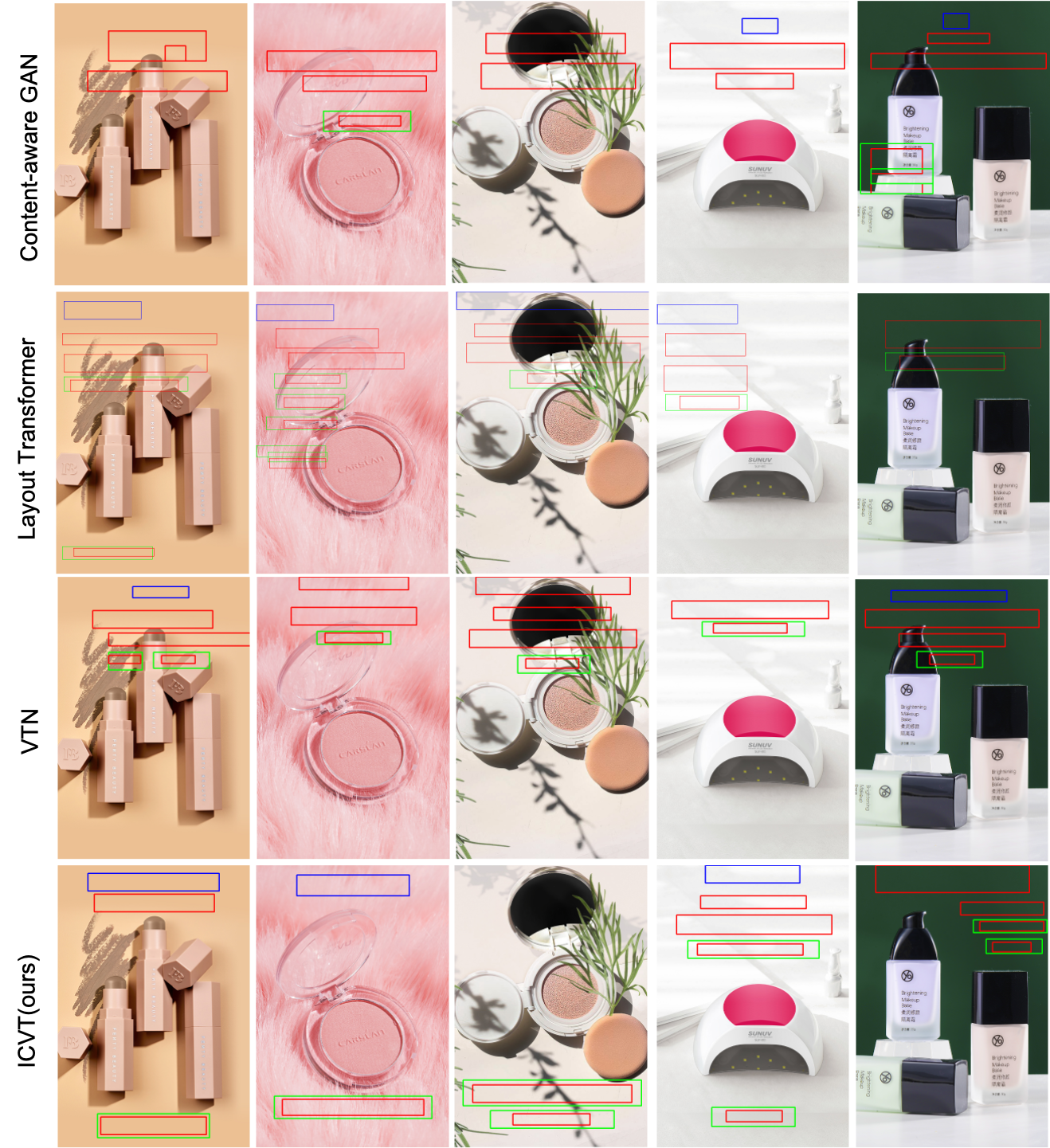}
      \caption{Comparison with content-aware GAN~\cite{contentGAN} (1st row), VTN~\cite{vtn} (2nd row), and Layout Transformer~\cite{layoutTransformer} (3rd row). Our method (4th row) produces bounding boxes around the salient object of the image without occlusion.}
      \label{fig:compare}
  \end{figure}

  As shown in Figure \ref{fig:compare}, We compare the generated results of our ICVT model with those of content-aware GAN~\cite{contentGAN}, VTN~\cite{vtn}, and Layout Transformer~\cite{layoutTransformer}. Although VTN generates high-quality layouts, they severely occlude salient objects of the image. The layouts generated by our ICVT model seldom occlude salient objects.
  \subsubsection{\textbf{Diversity and fidelity}}
\begin{figure}[htbp]
      \centering
      \includegraphics[width=\linewidth]{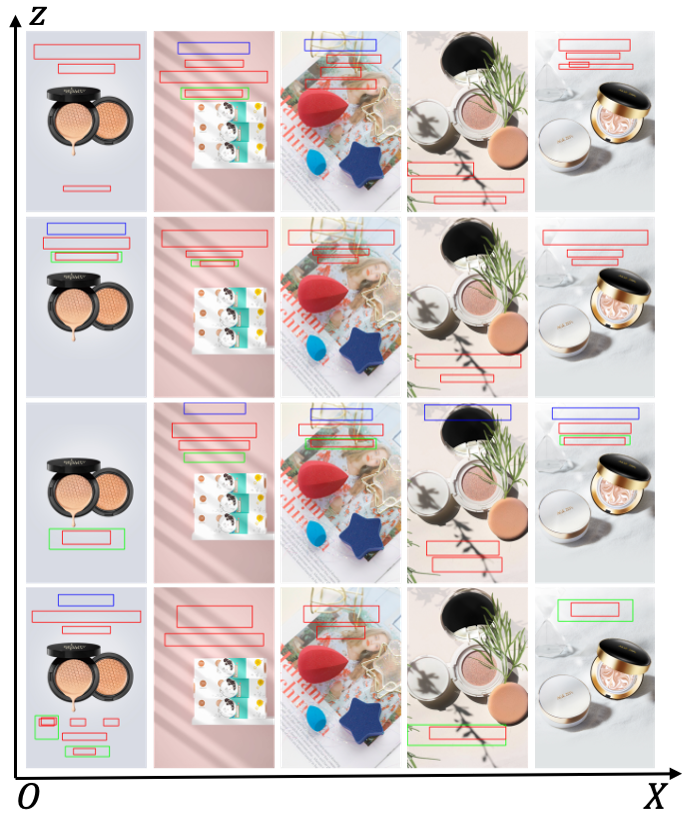}
      \caption{Diversity and fidelity. $\mathbf{X}$ refers to the conditional image and $\mathbf{z}$ refers to the latent vector. Each layout is an instance sampled from $P(\mathbf{Y}\mid \mathbf{z}, \mathbf{X})$.}
      \label{fig:diversity}
  \end{figure}

  Diversity and fidelity are two basic requirements for the task of image-conditioned layout generation. We need to generate diverse layouts while meeting the requirements of fidelity. 
  The decoder should learn decoupled representations of the two input variables, latent vector $\mathbf{z}$ and conditional image $\mathbf{X}$. 
  Therefore, we test our model by independently changing the above two variables respectively.
  As shown in Figure \ref{fig:diversity}, we change the conditional image $\mathbf{X}$ with latent vector $\mathbf{z}$ fixed and change $\mathbf{z}$ with $\mathbf{X}$ fixed. We find that with latent vector fixed(in the same row), the model still keeps fidelity under different image conditions. Similarly, with the conditional image fixed (in the same column), the model generates diverse layouts with different $\mathbf{z}$. 

\section{Conclusion}
  In this paper, we define and formulate the task of image-conditioned layout generation. We propose a novel transformer-based conditional variational autoencoder for image-conditioned layout generation, and construct a large-scale advertisement poster design layout dataset. We first parameterize layouts and extract information. Then we apply a transformer-based CVAE for jointly modeling the relationship between layouts and image. To alleviate the domain gap, we further design a geometry alignment module for enhancing spatial relationships. Finally, we perform extensive quantitative and qualitative experiments to demonstrate the effectiveness of our model.

\section*{Acknowledgements}
  This work is supported by Alibaba Group through Alibaba Innovation Research Program, the National Nature Science Foundation of China (62121002, 62022076, U1936210), the Fundamental Research Funds for the Central Universities under Grant WK3480000011, the Youth Innovation Promotion Association Chinese Academy of Sciences (Y2021122), the China Postdoctoral Science Foundation 2021M703081, and the Fundamental Research Funds for the Central Universities WK2100000026.

\clearpage
\balance
\bibliographystyle{ACM-Reference-Format}
\bibliography{sample-sigconf}

\clearpage
\appendix

\section{Dataset Analysis}\label{sec:data}
\begin{figure*}[htbp]
    \centering
    \includegraphics[width=\textwidth]{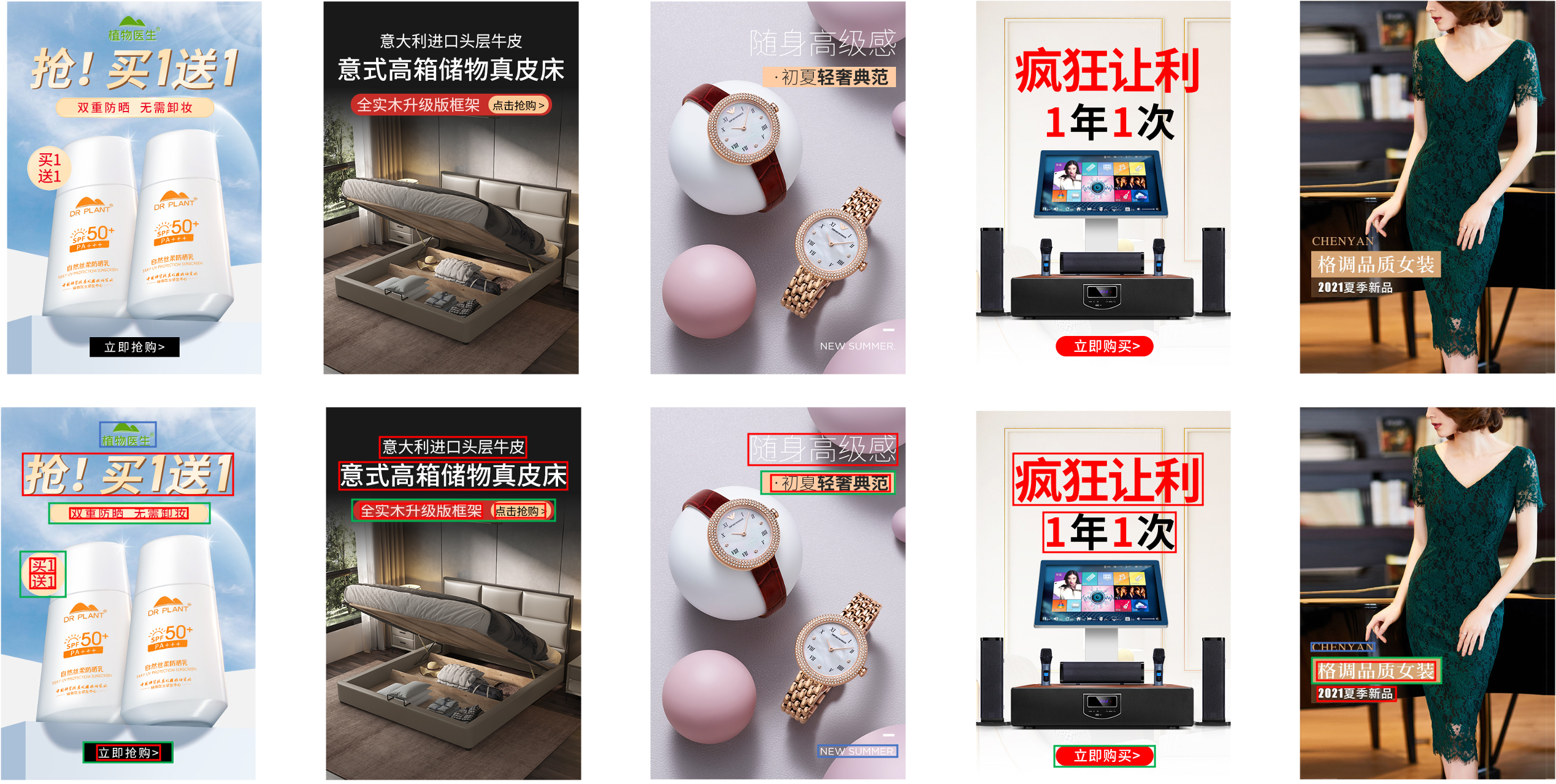}
    \caption{Showcase for the proposed dataset.}
    \label{fig:showcase}
\end{figure*}

The large-scale advertisement poster layout design dataset consists of 117,624 poster images designed by professional designers, annotated with rich in-image layout annotations. As shown in Figure~\ref{fig:showcase}, we present some samples of the dataset. Layout elements including texts, text substrates, and logos are annotated with red, green, and blue bounding boxes, respectively. Besides, we use a saliency map to 
indicate the position of the product subject, which guides the model to place layout elements on the non-invasive area in a semantically coherent manner.

\begin{figure*}[h]
    \centering
    \includegraphics[width=\textwidth]{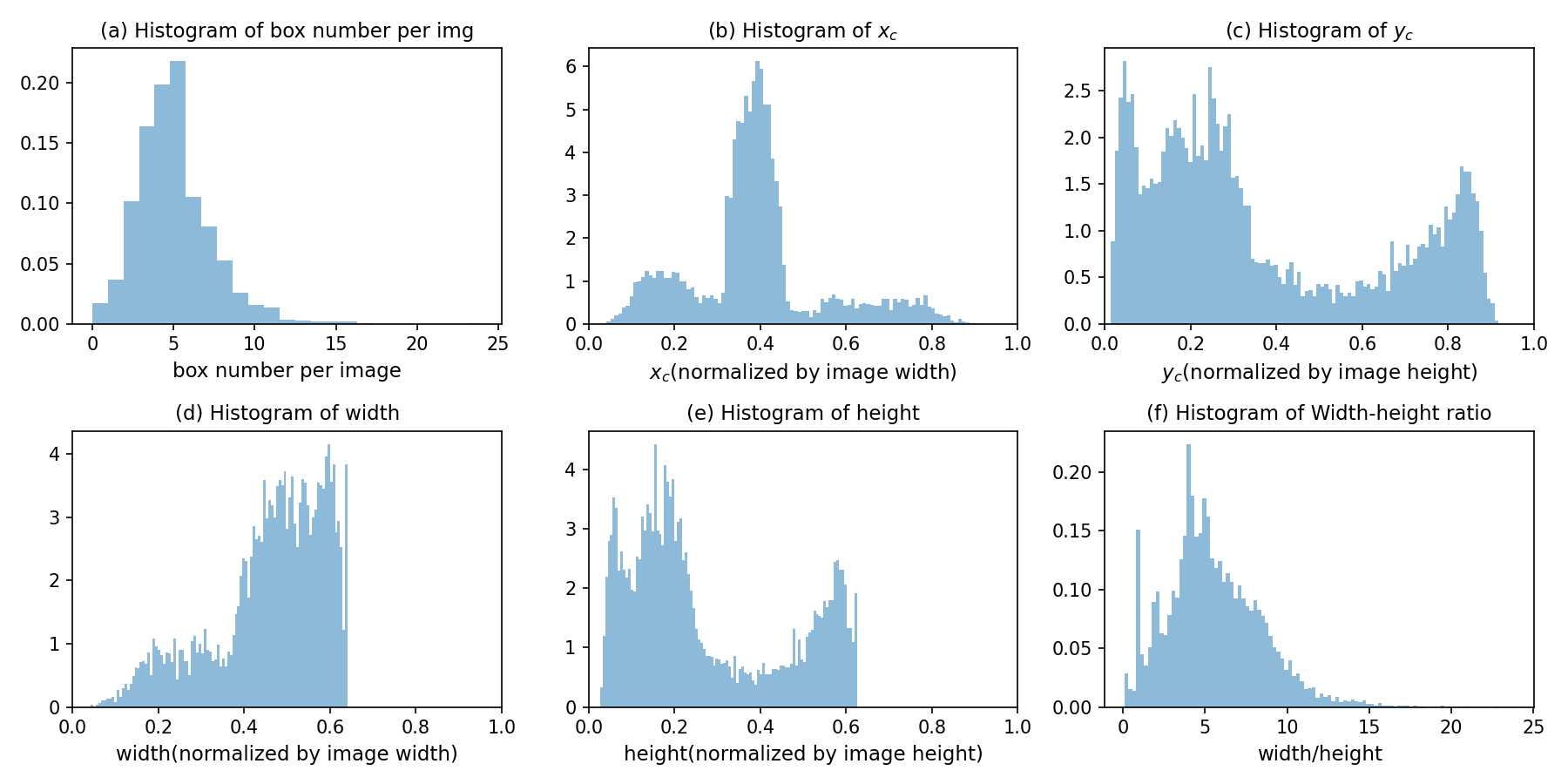}
    \caption{Layout bounding box distribution. Image (a) shows the distribution of bounding box numbers per image. In most cases, there are around 5 bounding boxes in an image. Image (b) (c) shows the coordinates of the center point of the bounding box. Most bounding boxes tend to be at the top and bottom of the image. Image (d) (e) shows the width and height of bounding boxes. As we can see, the width of most bounding boxes is 0.6 times the image width and the height is 0.2 times the image height. Image (f) shows the width-height ratio of bounding boxes, with the most probable value of around 5.}
    \label{fig:dist}
\end{figure*}

Furthermore, we analyze the dataset from different aspects, such as the number of bounding boxes per image, the coordinates of the center of bounding boxes, the width, and height of bounding boxes, and the width-height ratio, as shown in Figure~\ref{fig:dist}.

\section{Implementation Details}\label{sec: details}
We first provide a supplementary description of the implementation of our ICVT model. Then, considering that we reproduce the content-aware GAN~\cite{contentGAN}, Layout Transformer~\cite{layoutTransformer}, and VTN~\cite{vtn} to train them on our dataset, we describe the details of our implementation.

In our method, the model takes a masked image and corresponding layout design (sequence of bounding boxes) as input and reconstructs the layout conditioned on the image. We order layout elements according to their positions, from top to bottom and from left to right. We find that an ordered sequence is helpful for the model performance, although the layout elements can also be represented as an unordered set. For the input image, we resize the original product image into the size of $480\times 704$ and multiply it with a binary saliency mask, which highlights the product subject to guide the layout generation.

We implement the content-aware GAN~\cite{contentGAN} model with their open source code and downsample our input image into their proposed size of $45\times 60$. A larger resolution leads to a failure of training in our practice. We implement Layout Transformer~\cite{layoutTransformer} with their open source code. We implement VTN~\cite{vtn} following the description in their paper. The model dimension is set to 480 which differs from the dimension of 512 in their original VTN model for a fair comparison.

\section{Supplementary Experimental Results}\label{sec:results}
We present more uncurated random experimental results to fully evaluate the model performance.

\subsection{Layout Generation and Completion}
\begin{figure*}[p!]
    \centering
    \includegraphics[width=\textwidth]{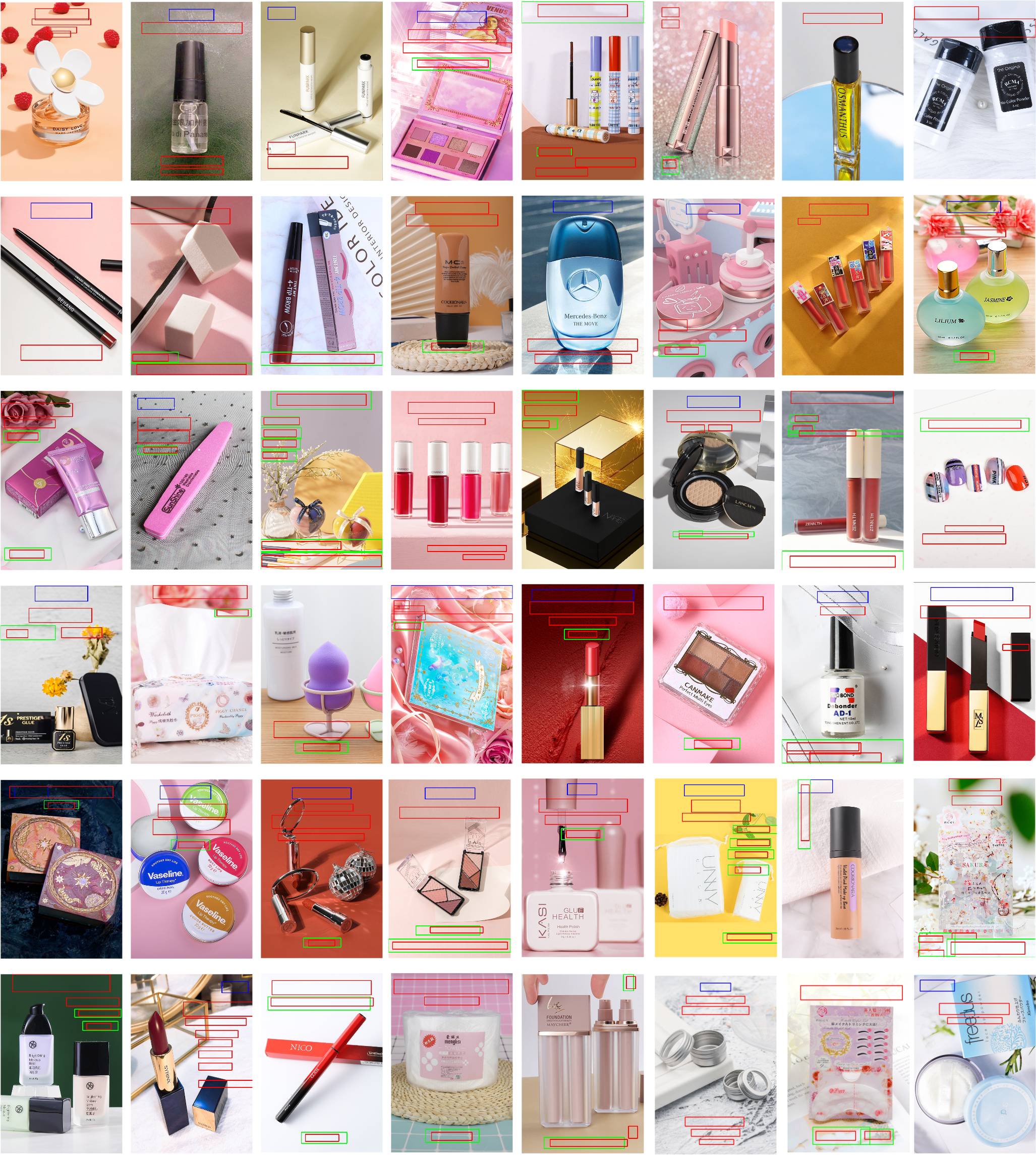}
    \caption{Layout Generation. We present uncurated random samples from the generation results of our ICVT model. In most cases, our model can generate proper layout design in the non-invasive area of the product image. Also, there are bad cases with problem of overlap and alignment.}
    \label{fig:gen_sup}
\end{figure*}

\begin{figure*}[p!]
    \centering
    \includegraphics[width=\textwidth]{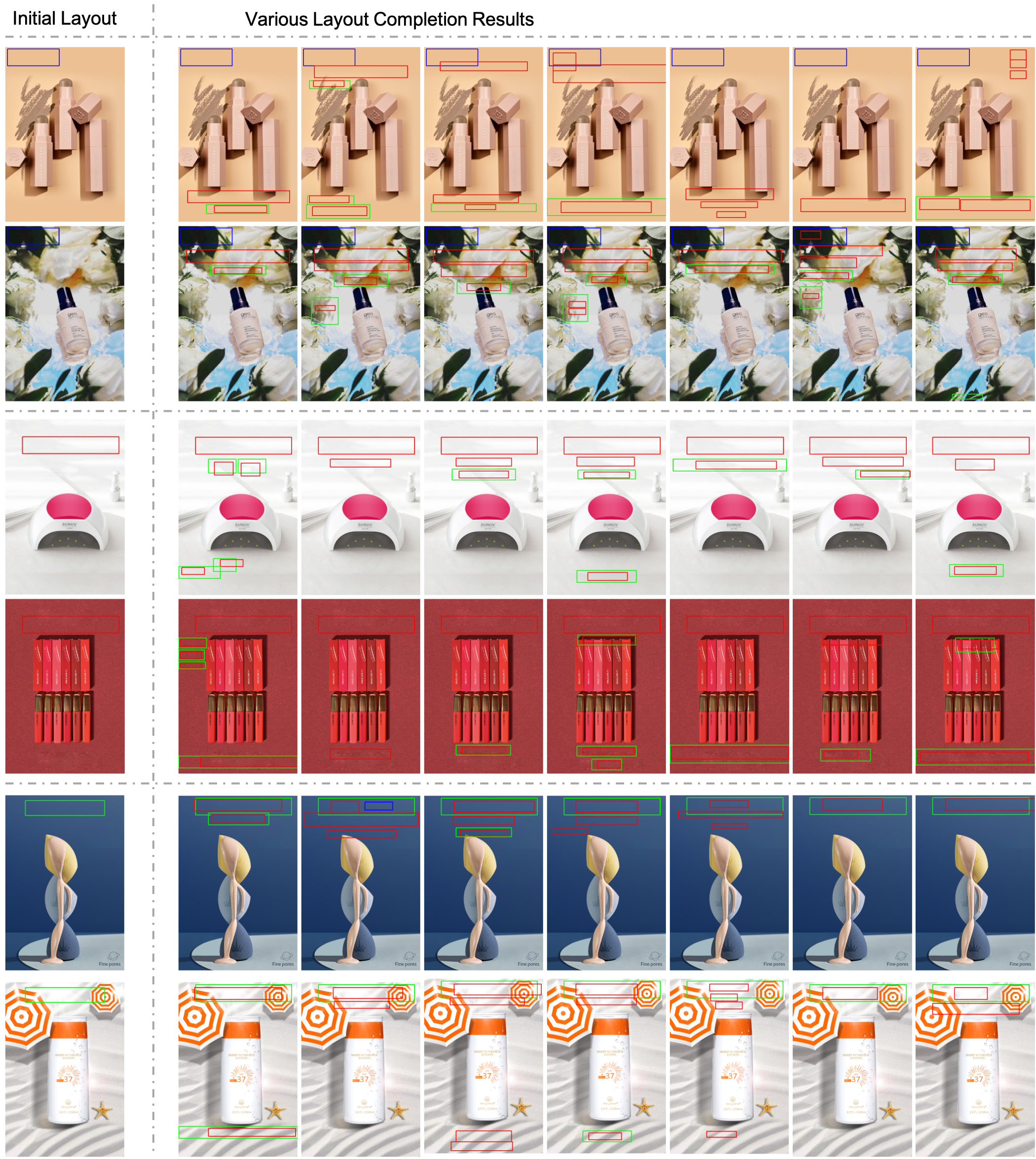}
    \caption{Layout Completion. The first column contains three different initial layouts of logos, texts, and text substrates. We generate various layout completions for each of them. In most cases, our model generates proper layouts conditioned on both the initial layout and image.}
    \label{fig:completion_sup}
\end{figure*}

Samples of layout generation and layout completion are shown in Figure~\ref{fig:gen_sup} and Figure~\ref{fig:completion_sup}. For layout completion, we take three different kinds of layout elements including logos, texts, and substrates as initial layouts and generate various layout completions. We find that our model can generate proper layouts conditioned both on the initial layout and images in most cases. Also, there are a few failure cases that produce overlapped layout elements.

\subsection{Comparison and User Study}
\begin{figure*}[p!]
    \centering
    \includegraphics[width=\textwidth]{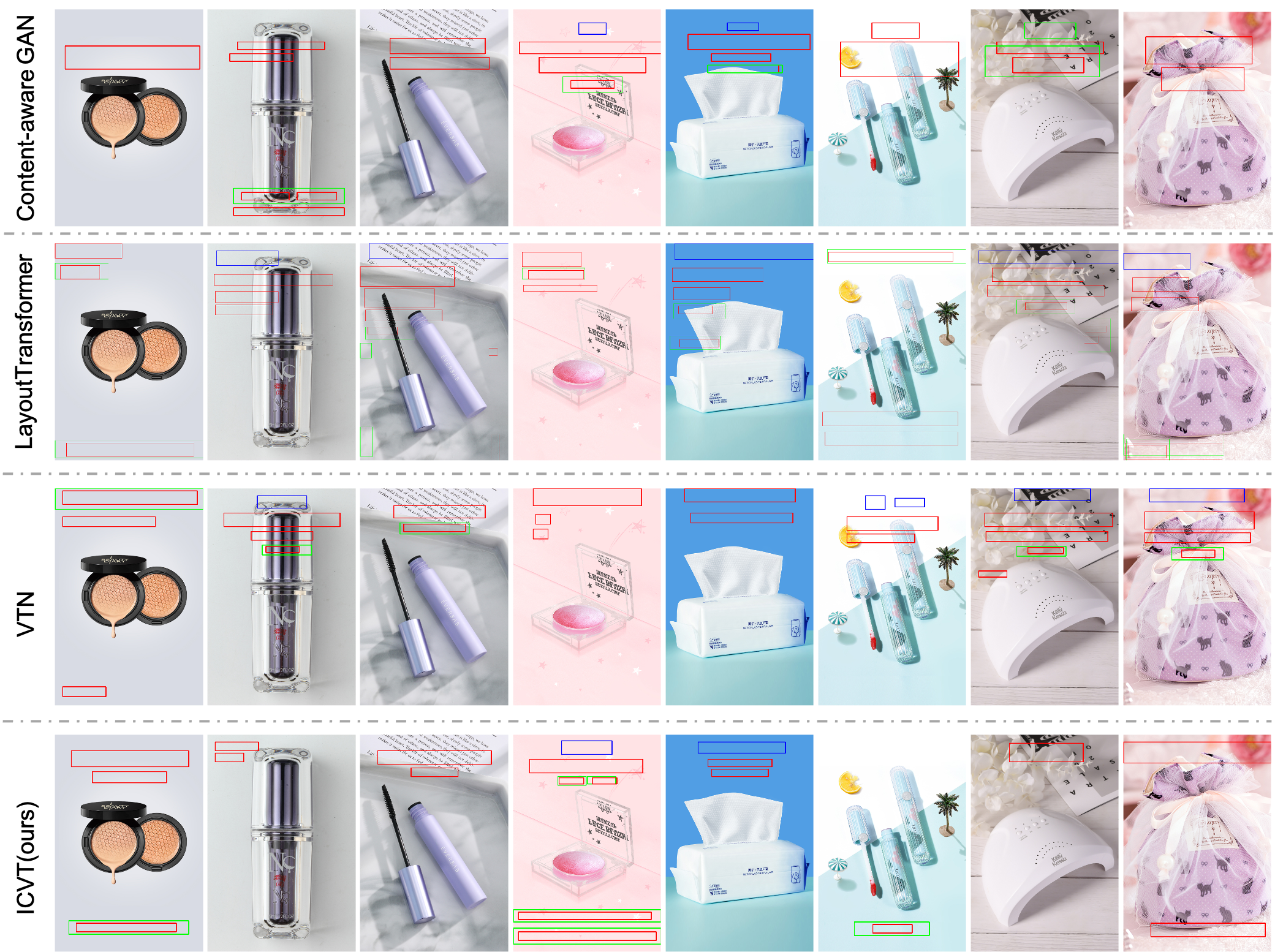}
    \caption{Comparison. We compare our model with the other three existing methods to demonstrate that our model generates a better in-image layout design, with fewer occlusion with the product subject.}
    \label{fig:compare_sup}
\end{figure*}

We also provide more results comparing our model with content-aware GAN~\cite{contentGAN}, LayoutTransformer~\cite{layoutTransformer}, and VTN~\cite{vtn}. As shown in Figure~\ref{fig:compare_sup}, we find that the occlusion problem in content-aware GAN\cite{layoutGAN}, LayoutTransformer~\cite{layoutTransformer}, and VTN~\cite{vtn} is much more serious, while most layout elements generated by our model are surrounding the product subject.

\begin{figure}[!h]
    \centering
    \includegraphics[width=\linewidth]{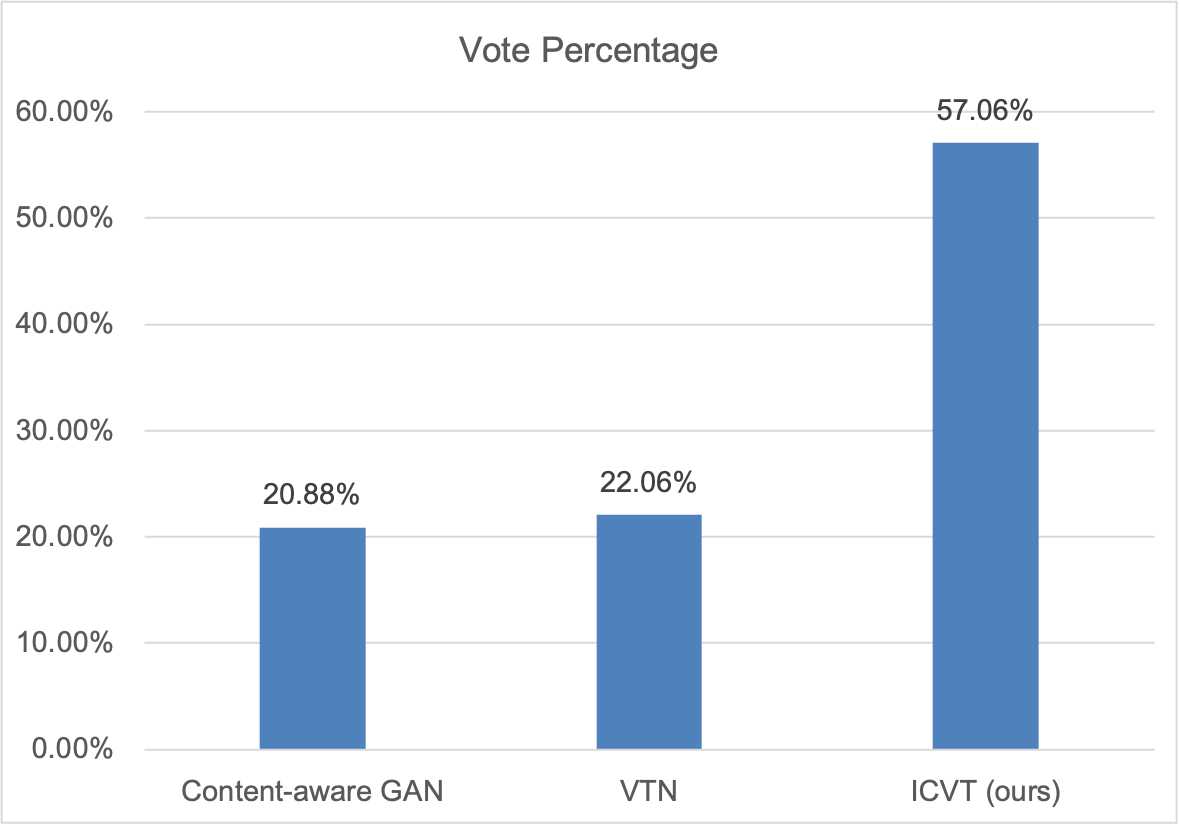}
    \caption{User Study. The histogram presents the percentage of times that the design results are voted to be the best.}
    \label{fig:user}
\end{figure}

Furthermore, we perform a user study to compare the design quality of different models. We randomly choose 20 images and 3 design results from different models for each conditional image. Subsequently, we let 51 volunteers vote for the best design for each image. After that, we statisticize the percentage of voting for different models. As shown in Figure~\ref{fig:user}, our model is the best.

\subsection{Diversity and Fidelity}
\begin{figure*}[p!]
    \centering
    \includegraphics[width=0.8\textwidth]{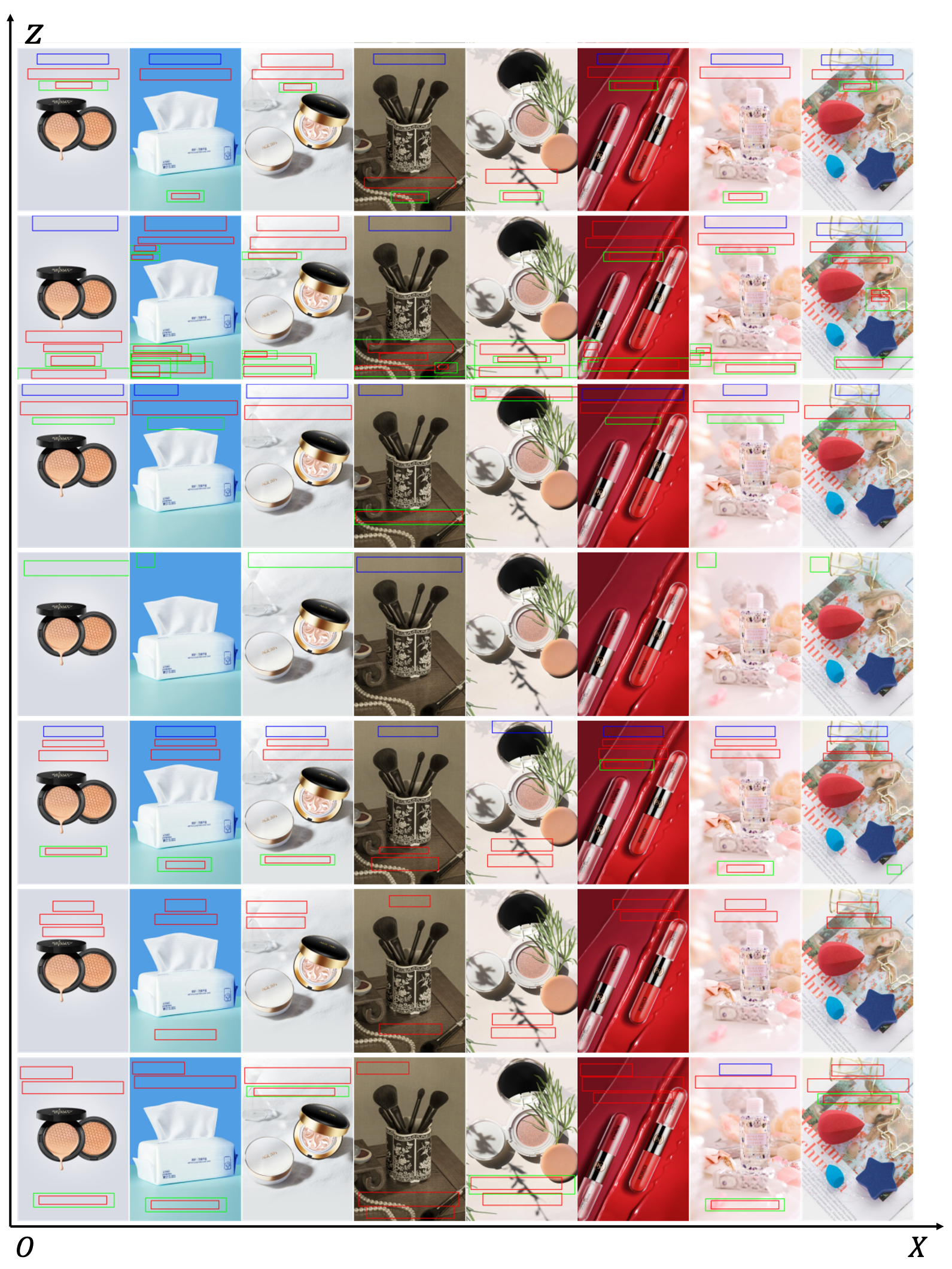}
    \caption{Diversity and Fidelity. The problem can be formulated as $\arg\max_{\mathbf{Y}} P(\mathbf{Y}\mid \mathbf{z}, \mathbf{X})$, where the prediction is determined by $\mathbf{z}$ and $\mathbf{X}$. We build a coordinate system of $\mathbf{X}\mathbf{O}\mathbf{z}$ to show the predictions given different $\mathbf{z}$ and $\mathbf{X}$. By fixing $\mathbf{X}$ and changing $\mathbf{z}$, we demonstrate the model diversity. By fixing $\mathbf{z}$ and changing $\mathbf{X}$, we demonstrate the model fidelity.}
    \label{fig:diversity_sup}
\end{figure*}

Diversity and fidelity are two basic aspects to measure generative models. To thoroughly exploit the performance of our model, we first formulate our problem as a conditional maximum likelihood estimation problem as follows:
\begin{equation}
    \hat{\mathbf{Y}}=\arg\max_{\mathbf{Y}} P(\mathbf{Y}\mid \mathbf{z}, \mathbf{X})
    \label{eq:p}
\end{equation}
where $\mathbf{Y}$,$\mathbf{z}$,$\mathbf{X}$ represent layout elements, latent vector, and conditional image, respectively. According to Equation~\ref{eq:p}, the prediction $\hat{\mathbf{Y}}$ is determined by $\mathbf{z}$ and $\mathbf{X}$, which represent the latent vector and conditional image, respectively. We evaluate our model by sampling in the 2-dimensional parameter space spanned by $\mathbf{z}$ and $\mathbf{X}$.
As shown in Figure~\ref{fig:diversity_sup}, we change the latent vector $\mathbf{z}$ in the direction of the vertical axis to show the diversity on the same image. Then, we fix latent vector $\mathbf{z}$ and change the conditional image in the direction of the horizontal axis to show the fidelity in different conditional images. Experimental results show that our model shows good performance in diversity and fidelity.

\end{document}